\documentclass[10pt,twocolumn,letterpaper]{article}

\usepackage{cvpr}
\usepackage{times}
\usepackage{epsfig}
\usepackage{graphicx}
\usepackage{amsmath}
\usepackage{amssymb}
\usepackage{booktabs}
\usepackage{subfigure}
\usepackage{graphics}
\usepackage{grffile}
\usepackage{tabularx}
\usepackage{multirow}

\usepackage[breaklinks=true,bookmarks=false]{hyperref}

\cvprfinalcopy 

\def\cvprPaperID{3287} 

\ifcvprfinal\fi
\begin{document}
\graphicspath{{pictures/}}
\title{Distilling Object Detectors with Fine-grained Feature Imitation}

\author{\normalsize{Tao~Wang$^1$} \qquad \quad  \normalsize{Li~Yuan$^1$}  \quad  \qquad \normalsize{Xiaopeng~Zhang$^{1,2}$}  \quad  \qquad \normalsize{Jiashi~Feng$^1$}\\
	\small{$^{1}$Department of Electrical and Computer Engineering, National University of Singapore, Singapore} \\
	\small{$^{2}$Huawei Noah's Ark Lab, Shanghai, China} \\
	{\small \tt twangnh@gmail.com}  \ \  {\small \tt ylustcnus@gmail.com}  \ \ {\small\tt zhangxiaopeng12@huawei.com} \ \ {\small\tt elefjia@nus.edu.sg}
}

\maketitle
\thispagestyle{empty}


\begin{abstract}	

State-of-the-art CNN based recognition models are often computationally prohibitive to deploy on low-end devices. A promising high level approach tackling this limitation is \emph{knowledge distillation}, which let small student model mimic cumbersome teacher model's output to get improved generalization. However, related methods mainly focus on simple task of classification while do not consider complex tasks like object detection. We show applying the vanilla knowledge distillation to detection model gets minor gain.  To address the challenge of distilling knowledge in detection model, we propose a \emph{fine-grained feature imitation} method exploiting the cross-location discrepancy of feature response. Our intuition is that detectors care more about local near object regions. Thus the discrepancy of feature response on the near object anchor locations reveals important information of how teacher model tends to generalize. We design a novel mechanism to estimate those locations and let student model imitate the teacher on them to get enhanced performance. We first validate the idea on a developed lightweight toy detector which carries simplest notion of current state-of-the-art anchor based detection models on challenging KITTI dataset, our method generates up to 15\% boost of mAP for the student model compared to the non-imitated counterpart. We then extensively evaluate the method with Faster R-CNN model under various scenarios with common object detection benchmark of Pascal VOC and COCO, imitation alleviates up to 74\% performance drop of student model compared to teacher. Codes released at \url{https://github.com/twangnh/Distilling-Object-Detectors}
\end{abstract}

\begin{figure}[!t]
	\centering
	\includegraphics[width=3.2in, height=1.8in]{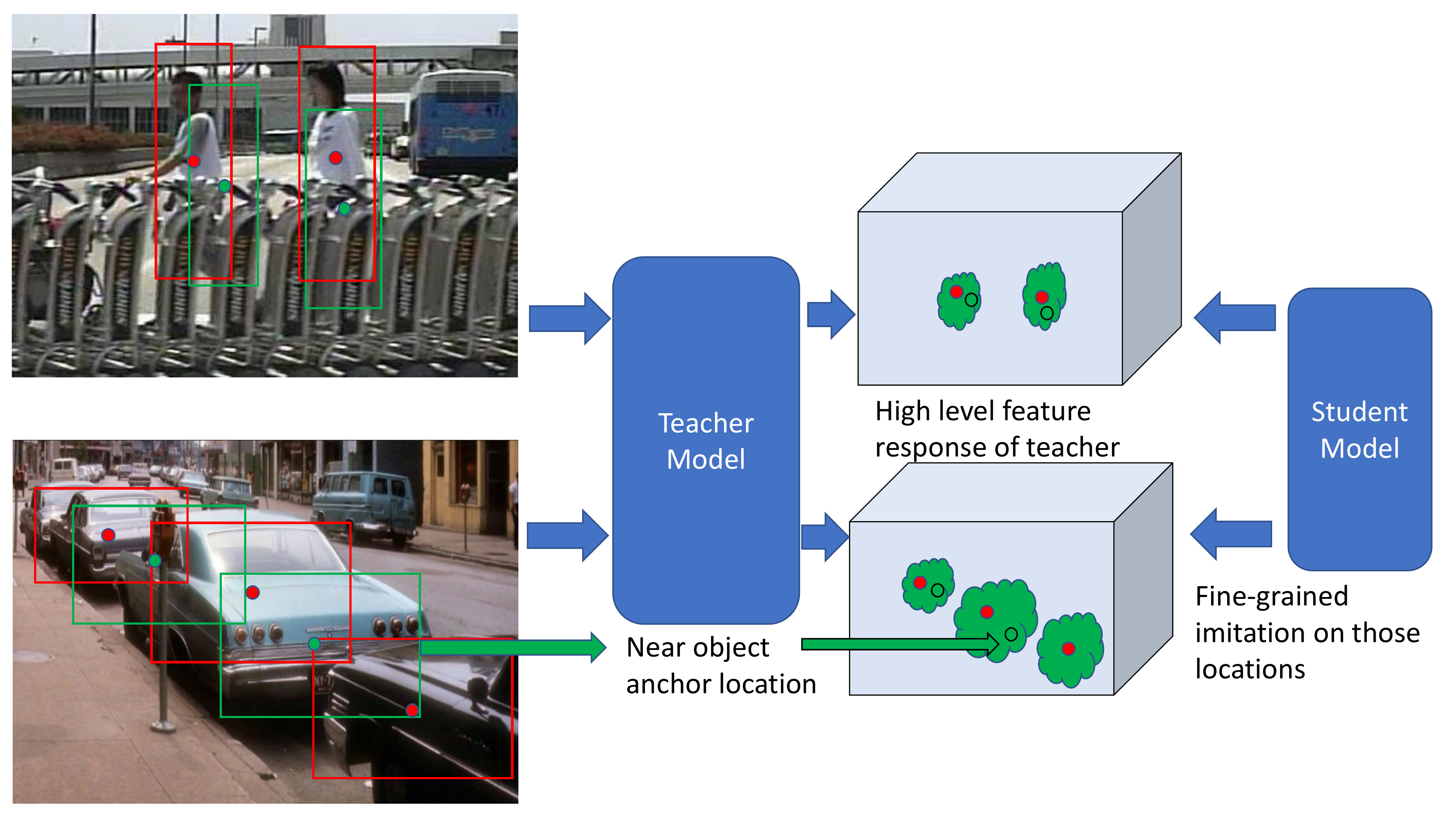}
	\caption{Illustration on principle of the proposed method. Red and green bounding boxes on the left two images are selected prior anchor boxes on corresponding locations. The red anchors have the largest overlap with ground truth bounding boxes and the green ones indicate near object samples. 
	The motivation is that the discrepancy of feature response on near object anchor locations reveals how a learned teacher model tends to generalize (\emph{e.g.}, how the teacher responses on those intersections of crowed objects compared to on-object locations reflects how it separates and detects those crowded instances). Our method thus first locates these knowledge-dense locations and let the student model imitate teacher's high-level feature responses on them.}
	\label{intuition}
	\vspace{-10pt}
\end{figure}

\section{Introduction}
\label{intro}
Object detection has benefited a lot from recent advances of deep CNN architectures. However state-of-art detectors are cumbersome to deploy on low computation devices. Previous works mainly focus on Quantization~\cite{Gysel2016HardwareorientedAO, Han2015LearningBW, Wu2016QuantizedCN, rastegari2016xnor} which efficiently reduces computation and model size, and network pruning~\cite{Han2015LearningBW, Han2015DeepCC, alvarez2016learning, wen2016learning} that prunes redundant connections in large models. However these approaches may require dedicated hardware or software customization to get practical speedup.

A promising high level method to directly learn compact models end-to-end is \emph{knowledge distillation}~\cite{Hinton2015DistillingTK}. A student model learns the behavior of a stronger teacher network to get enhanced generalization.
However, prior works on knowledge distillation~\cite{Hinton2015DistillingTK,romero2014fitnets, zagoruyko2016paying, chen2017darkrank, huang2017like} are mostly devoted to classification and rarely consider object detection. A detection model may only involve a few classes, with which much less knowledge can be distilled from inter-class similarity of teacher's softened outputs. Also, detection requires reliable localization in addition to classification, vanilla distillation can not be applied for distilling localization knowledge. Besides, the extreme imbalance of foreground and background instances also makes bounding box annotations less voluminous.
We find that merely adding distillation loss only gives minor boost for student (ref. Sec. \ref{exp_imitation_toy_detector}). 

Similar to knowledge distillation, hint learning~\cite{romero2014fitnets} improves student models by minimizing the discrepancy of full high level features of the teacher and student models. But we find that directly applying hint learning on detection model hurts performance (ref. Sec. \ref{exp_imitation_toy_detector}). The intuition is that detectors care more about local regions that overlap with ground truth objects while classification models pay more attention to global context. So directly doing full feature imitation would unavoidably introduces large amount of noise from uncared areas, especially for object detection where background instances are overwhelming and diverse. 

Recall in knowledge distillation, relative probabilities on different classes indeed tell a lot about how the teacher model tends to generalize. Similarly, since detectors care more about local object regions, the discrepancy of feature response on close anchor locations near the object also conveys important information about how a complex detection model detects the object instances. Aiming to utilize this \emph{inter-location discrepancy} for distilling knowledge in object detector, we develop a novel mechanism exploiting ground truth bounding boxes and anchor priors to effectively estimate those informative near object anchor locations, then make student model imitate teacher on them, as shown in Figure~\ref{intuition}. 

We term this method as \emph{fine-grained feature imitation}. Our method effectively addresses the above mentioned challenge: 1) We do not rely on softened output of teacher model as in vanilla knowledge distillation of classification model, but depends on a \emph{inter-location} discrepancy of teacher's high level feature response. 2) Fine-grained feature imitation before classification and localization heads improves both sub-tasks. We show in Sec~\ref{exp_imitation_gain_qualitative} and Sec~\ref{exp_imitation_gain_error_perspective} that our method effectively enhanced the student model's ability on class discrimination and localization. 3) Our method avoids those noisy less informative background area which leads to degraded performance of full feature imitation, study of the \emph{per-channel variance} on high level feature maps in Sec~\ref{per_channel_variance} validates this intuition.

To validate our method, we first experiment on a developed lightweight toy detector that carries main principle of current state-of-the-art anchor based detection models. Applying the method to this  lightweight architecture, we can produce much smaller model with up to 15\% boost of mAP compared to the non-imitated counterpart. We then perform extensive experiments on the state-fo-the-art Faster R-CNN model under various scenarios including imitation over shallow student, halved student and multi-layer imitation, on the widely used common object detection benchmarks of PASCAL VOC~\cite{everingham2010pascal} and MSCOCO~\cite{lin2014microsoft}. The experiments demonstrate the broad applicability and superior performance of our proposed method. 

\section{Related Works}

\paragraph{Object detection} 
Recently with the development of deep CNN model for image classification task, various approaches~\cite{girshick2014rich, girshick2015fast, ren2015faster, cai2017cascade, redmon2016you, redmon2017yolo9000, liu2016ssd, lin2018focal} are proposed for object detection which significantly outperform traditional methods. The line of works are pioneered by R-CNN~\cite{girshick2014rich} that extracts and classifies each region of interest (ROI) to detect objects. ~\cite{girshick2015fast, ren2015faster} extend and improve the framework for improved performance. One-stage detectors~\cite{redmon2016you, liu2016ssd} are proposed driven by the requirement of real time inference. Similarly we design the lightweight detector partly for implementation on mobile devices.

\vspace{-4mm}
\paragraph{Knowledge distillation} Following the seminal work~\cite{hinton2015distilling}, various knowledge distillation approaches were proposed~\cite{romero2014fitnets,zagoruyko2016paying,chen2017darkrank, huang2017like}. Hint learning~\cite{romero2014fitnets} explores an alternative way for distillation, where the supervision from teacher models comes from high level features.   \cite{zagoruyko2016paying} proposed to 
force the student model to mimic the teacher model on the features specified by an attention map. 
\cite{chen2017darkrank} proposed to exploit relationship between different samples, and utilizes cross sample similarities to improve distillation. \cite{huang2017like} formalizes distillation as a distribution matching problem to optimize the student model. 
A few recent works explored distillation approach for compressing detection models. \cite{chen2017learning} tried adding both full feature imitation and specific distillation loss on detection heads, but we find full feature imitation brings degraded performance for student model and it is unclear how to deal with region proposal ~\cite{Girshick2015FastR} inconsistency between teacher and student when performing the distillation.~\cite{li2017mimicking} proposed to only transfer knowledge under the area of proposals, but the mimicking regions depend on the output of model itself and it is not applicable for one-stage detector.

\vspace{-4mm}
\paragraph{Model acceleration} To speed up  deep neural network model without losing accuracy,  quantization~\cite{zhou2017incremental, rastegari2016xnor, wu2016quantized, Gysel2016HardwareorientedAO, Han2015LearningBW, Wu2016QuantizedCN}  uses low-precision model parameter representation. Connection pruning or weight sparsifying~\cite{Han2015LearningBW, Han2015DeepCC, park2016faster} prune redundant connections in large models. However, these approaches require specific hardware  or software customization to get practical speedup. For example, weight pruning needs support of sparse computations and quantization relies on low-bit operations. Some prior works~\cite{li2016pruning,luo2017thinet, anisimov2017towards} propose to do channel level pruning. But when pruning ratio is higher, those methods unavoidably hurt performance significantly. Some works employ low rank approximation to large layers~\cite{tai2015convolutional, wen2017coordinating}. But the actual speedup are usually much less than theoretical values.

\begin{figure}[!t]
	\centering
	\includegraphics[width=1\linewidth]{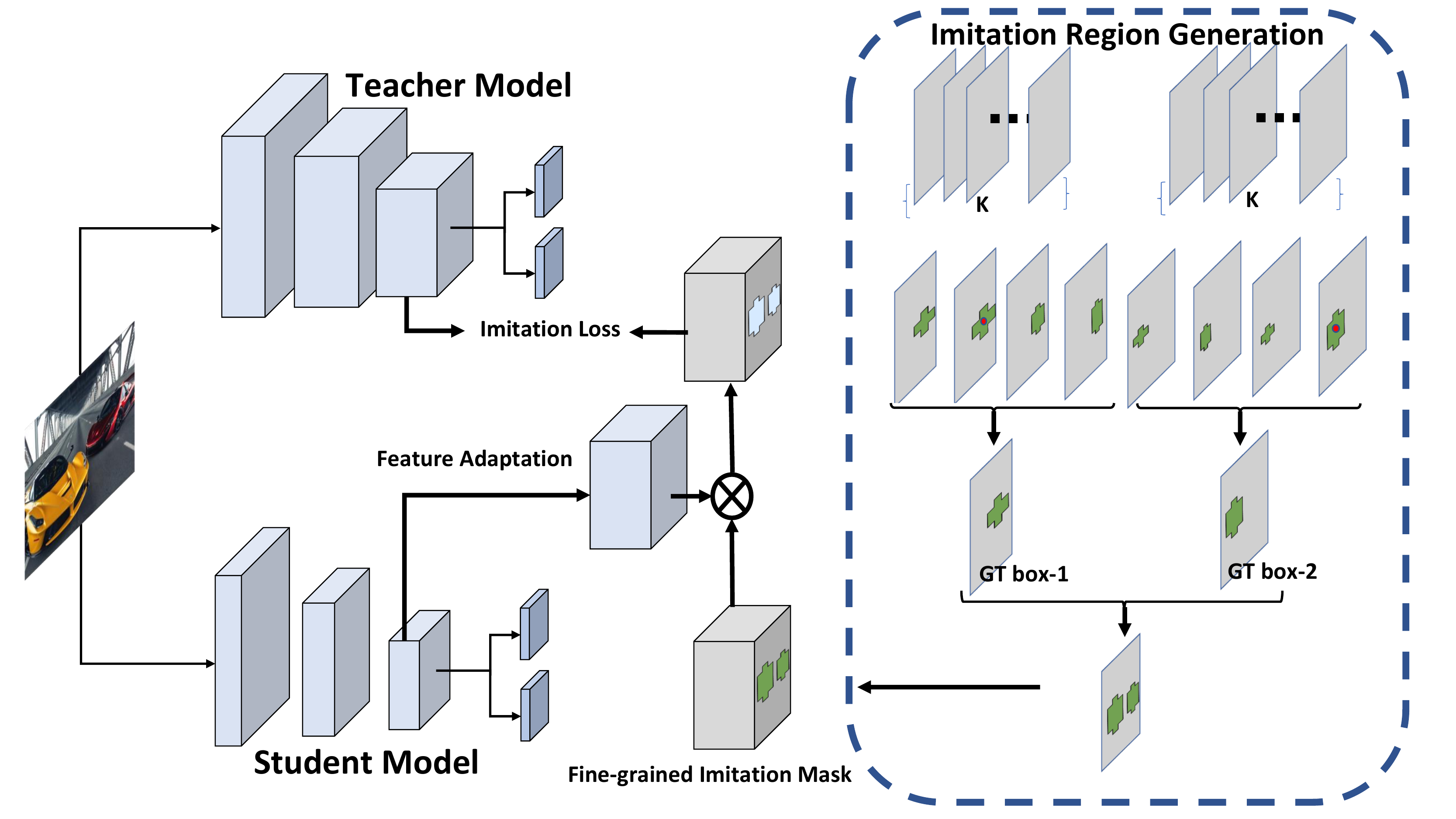}
	\caption{Illustration of the proposed fine-grained feature imitation method. The student detector is trained by both ground truth supervision and imitating  teacher's feature response on close object anchor locations. The feature-adaptation layer makes  student's guided feature layer compatible with the teacher. To identify informative locations, we iteratively calculate IOU map of each groundtruth bounding box with anchor priors, filter and combine candidates, and generate the final imitation mask, ref. to Sec.~\ref{sec_imitation_method} for details. 
	}
	\label{framework}
	\vspace{-4mm}
\end{figure}

\section{Method}


In this work, we developed a simple to implement fine-grained feature imitation method utilizing inter-location discrepancy of teacher's feature response on near object anchor locations for distilling the knowledge in cumbersome detection models. Our Intuition is that the discrepancy of feature response on the near object anchor locations reveals important information of how large detector tends to generalize, with which learned knowledge can be distilled. Specifically, we propose a novel mechanism to estimate those anchor locations which forms fine-grained local feature regions close to object instances, and let a student model imitate teacher model's high level feature response on those regions to get enhanced performance. This intuitive method is general for current state-of-the-art anchor based detection models (\textit{e.g.}, Faster R-CNN~\cite{ren2015faster}, SSD~\cite{liu2016ssd}, YOLOV2~\cite{redmon2017yolo9000}), and is orthogonal to other model acceleration methods including network pruning and quantization.

\subsection{Imitation region estimation}
\label{sec_imitation_method}

As shown in Fig.~\ref{intuition}, the near object anchor locations form local feature region for each object. To formally define and study the local feature region,  we utilize ground truth bounding boxes and anchor priors to calculate those regions as a mask $I$ for each independent image, and control the size of regions by a thresholding factor $\psi$. In the following, with feature maps, we always refer to the  last features where anchor priors are defined on~\cite{ren2015faster}.

Specifically, as shown in Fig.~\ref{framework}, for each ground truth box, we compute the IOU between it and all anchors, which forms a $W\times H\times K$ IOU map $ m $. Here $W$ and $H$ denote  width and height of the feature map, and $K$ indicates the $K$ preset anchor boxes. Then we find the largest IOU value $ M=max(m) $, times the thresholding factor $\psi$ to obtain a filter threshold $ F=\psi*M $. With $ F $, we filter the IOU map to keep those larger then $ F $ locations and combine them with OR operation to get a $W\times H$ mask. Loop over all ground truth boxes and combine the masks, we get the final fine-grained imitation mask $ I $.

When $ \psi=0 $, the generated mask includes all locations on the feature map while no locations are kept when $ \psi=1 $. We can get varied imitation mask by varying $ \psi $. In all experiments, a constant $ \psi =0.5$ is used. We show $ \psi =0.5$ offers the best distillation performance in detailed ablation study (ref. to Sec~\ref{vary_psi}). The reason we do not use fixed value of $ F $ to filter the IOU map is that object size usually varies in a large range. Fixed threshold values would be biased for objects at certain scales and ratios (ref. Sec.~\ref{exp_imitation_toy_detector}).

\subsection{Fine-grained feature imitation}

In order to carry out imitation, we add a full convolution adaptation layer after corresponding student model before calculating distance metric between student and teacher's feature response, as shown in Figure~\ref{framework}.  We add the adaptation layer for two reasons: 1) The student feature's channel number may not be compatible with teacher model. The added layer can align the former to the later for calculating distance metric. 2) We find even when student and teacher have compatible features, forcing student to approximate teacher feature directly leads to minor gains compared to the adapted counterpart.

We now introduce the feature imitation details. Define $s$ as student model's guided feature map and  $t$ as corresponding teacher's feature map. For each near object anchor location $ (i, j) $ on the feature map of width $ W $ and height $ H $, we train student model to minimize  the following objective:
\begin{equation}
\begin{aligned}
l =
\sum_{c=1}^C(f_{\mathrm{adap}}(s)_{ijc}-t_{ijc})^2,
\end{aligned}
\end{equation}
to learn the teacher detection model's knowledge. Together with all estimated near anchor location(the imitation mask $ I $), the distillation objective is to minimize:
\begin{equation}
\begin{aligned}
&L_{imitation} = \frac{1}{2N_{p}}\sum_{i=1}^W\sum_{j=1}^H\sum_{c=1}^CI_{ij}(f_{\mathrm{adap}}(s)_{ijc}-t_{ijc})^2, \\&\text{ where } N_{p} = \sum_{i=1}^W\sum_{j=1}^HI_{ij}.
\end{aligned}
\end{equation}
Here $I$ is the imitation mask, $N_{p}$ is the number of positive points in the mask,  $f_{\mathrm{adap}}(\cdot)$ is the adaptation function. Then the overall training loss of a student model is:
\begin{equation}
L = L_{gt} + \lambda L_{imitation},
\end{equation}
where $L_{gt}$ is the detection training loss and $\lambda$ is imitation loss weight balancing factor.

\begin{table*}[]
	\centering
	\renewcommand{\tabcolsep}{4.8pt}
	\small
	\begin{tabular}
		{@{\extracolsep{2pt}}cccccccccccccr@{}}
		\toprule 
		\multirow{2}{*}{Models}                                                        & \multirow{2}{*}{\begin{tabular}[c]{@{}c@{}}Flops/G\end{tabular}} & \multirow{2}{*}{\begin{tabular}[c]{@{}c@{}}Params/M\end{tabular}} & \multicolumn{3}{c}{\emph{car}}                                                                                                                                                                                     & \multicolumn{3}{c}{\emph{pedestrian}}                                                                                                                                                                           & \multicolumn{3}{c}{\emph{cyclist}}                                                                                                                                                                               & \multirow{2}{*}{mAP}                                              \\ \cmidrule{4-6} \cmidrule{7-9} \cmidrule{10-12}
		&                                                                     &                                                                      & Easy                                                                & Moderate                                                               & Hard                                                               & Easy                                                              & Moderate                                                               & Hard                                                              & Easy                                                               & Moderate                                                               & Hard                                                              &                                                                   \\ \midrule 
		1$\times$                                                                             & 5.1                                                                 & 1.6                                                                  & 84.56                                                               & 74.11                                                              & 65.64                                                              & 65.28                                                             & 55.95                                                             & 50.79                                                             & 70.39                                                              & 50.09                                                             & 46.88                                                             & 62.63                                                             \\ \midrule 
		\begin{tabular}[c]{@{}c@{}}0.5$\times$\\ 0.5$\times$-I\\ -\end{tabular}                      & \begin{tabular}[c]{@{}c@{}}1.5\\ 1.5\\ -\end{tabular}               & \begin{tabular}[c]{@{}c@{}}0.53\\ 0.53\\ -\end{tabular}              & \begin{tabular}[c]{@{}c@{}}76.39\\ 80.56\\ \textbf{+4.2}\end{tabular}        & \begin{tabular}[c]{@{}c@{}}68.35\\ 71.46\\ \textbf{+3.1}\end{tabular}       & \begin{tabular}[c]{@{}c@{}}59.74\\ 61.71\\ \textbf{+2.0}\end{tabular}       & \begin{tabular}[c]{@{}c@{}}63.69\\ 64.18\\ \textbf{+0.5}\end{tabular}      & \begin{tabular}[c]{@{}c@{}}54.34\\ 54.62\\ \textbf{+0.3}\end{tabular}      & \begin{tabular}[c]{@{}c@{}}49.58\\ 49.95\\ \textbf{+0.4}\end{tabular}      & \begin{tabular}[c]{@{}c@{}}64.52\\ 68.25\\ \textbf{+3.7}\end{tabular}       & \begin{tabular}[c]{@{}c@{}}43.67\\ 48.28\\ \textbf{+4.6}\end{tabular}      & \begin{tabular}[c]{@{}c@{}}41.57\\ 45.09\\ \textbf{+3.5}\end{tabular}      & \begin{tabular}[c]{@{}c@{}}57.98\\ 60.46\\ \textbf{+2.5}\end{tabular}      \\ \midrule 
		\begin{tabular}[c]{@{}c@{}}0.25$\times$\\ 0.25$\times$-I\\ -\end{tabular}                    & \begin{tabular}[c]{@{}c@{}}0.67\\ 0.67\\ -\end{tabular}             & \begin{tabular}[c]{@{}c@{}}0.21\\ 0.21\\ -\end{tabular}              & \begin{tabular}[c]{@{}c@{}}60.36\\ 74.26\\ \textbf{+13.9}\end{tabular}       & \begin{tabular}[c]{@{}c@{}}54.85\\ 61.63\\ \textbf{+6.8}\end{tabular}       & \begin{tabular}[c]{@{}c@{}}46.56\\ 53.94\\ \textbf{+7.4}\end{tabular}       & \begin{tabular}[c]{@{}c@{}}52.41\\ 59.80\\ \textbf{+7.4}\end{tabular}      & \begin{tabular}[c]{@{}c@{}}43.63\\ 50.15\\ \textbf{+6.5}\end{tabular}      & \begin{tabular}[c]{@{}c@{}}39.84\\ 46.28\\ \textbf{+6.4}\end{tabular}      & \begin{tabular}[c]{@{}c@{}}51.35\\ 54.64\\ \textbf{+3.3}\end{tabular}       & \begin{tabular}[c]{@{}c@{}}33.41\\ 38.13\\ \textbf{+4.7}\end{tabular}      & \begin{tabular}[c]{@{}c@{}}31.26\\ 34.84\\ \textbf{+3.6}\end{tabular}      & \begin{tabular}[c]{@{}c@{}}45.96\\ 52.63\\ \textbf{+6.7}\end{tabular}      \\ \midrule 
		\begin{tabular}[c]{@{}c@{}}0.25$\times$-F\\ 0.25$\times$-G\\ 0.25$\times$-D\\ 0.25$\times$-ID\end{tabular} & \begin{tabular}[c]{@{}c@{}}0.67\\ 0.67\\ 0.67\\ 0.67\end{tabular}   & \begin{tabular}[c]{@{}c@{}}0.21\\ 0.21\\ 0.21\\ 0.21\end{tabular}    & \begin{tabular}[c]{@{}c@{}}-12.9\\ +8.8\\ +3.5\\ +10.8\end{tabular} & \begin{tabular}[c]{@{}c@{}}-14.5\\ +2.3\\ +1.2\\ +5.8\end{tabular} & \begin{tabular}[c]{@{}c@{}}-11.3\\ +1.2\\ +1.3\\ +6.3\end{tabular} & \begin{tabular}[c]{@{}c@{}}-2.9\\ +3.1\\ +1.1\\ +6.2\end{tabular} & \begin{tabular}[c]{@{}c@{}}-1.9\\ +0.8\\ +0.8\\ +4.1\end{tabular} & \begin{tabular}[c]{@{}c@{}}-1.3\\ +2.4\\ +0.3\\ +3.6\end{tabular} & \begin{tabular}[c]{@{}c@{}}-16.7\\ -0.5\\ +0.2\\ +2.2\end{tabular} & \begin{tabular}[c]{@{}c@{}}-9.3\\ -0.1\\ -0.3\\ +4.7\end{tabular} & \begin{tabular}[c]{@{}c@{}}-9.4\\ -0.3\\ -0.1\\ +3.1\end{tabular} & \begin{tabular}[c]{@{}c@{}}-8.9\\ +2.0\\ +0.9\\ +5.2\end{tabular} \\ \bottomrule 
	\end{tabular}
	\label{exp_toy_detector}
	\caption{Imitation result on the toy detector and results of some comparing methods. $1\times $ is the base detector, $0.5\times $ and $0.25\times $ are directly pruned model trained with ground truth supervision, serving as baselines. -I means with additional proposed imitation loss, -F indicates with full feature imitation, -G means using directly scaled ground truth boxes as imitation region, -D means adding only vanilla distillation loss, -ID indicates the case that both proposed imitation loss and distillation loss are imposed.}
	\vspace{-3mm}
\end{table*}

\section{Experiments} \label{experiment_section}
To validate our method, we first perform experiments on a developed lightweight toy detector with the KITTI detection benchmark which contains three road object classes. We then further validate the method on state-of-the-art Faster R-CNN model under various network setting with widely used common object detection benchmarks. 
The toy detector carries simplest principle of state-of-the-art anchor based detection model, while the performance is not comparable to those cumbersome and multi-stage stage or multi-layer detection models, it can applied to mobile devices.
All quantitative results are evaluated in average precision (AP).

\subsection{Lightweight detector}
\label{sec_lightweight_detector}

We first present a manually designed lightweight detector for evaluating the performance enhancement of the proposed imitation method. This detector is based on the {Shufflenet}~\cite{shufflenet} which gives excellent classification performance with limited flops and parameters.  However, the Shufflenet architecture itself is dedicated for image classification. We find directly adapting it to detection produces terrible result. This is because each point on the top feature map has an equivalent stride of 32, leading to very coarse alignment of anchor boxes on the input images.  Moving to lower output layer with smaller stride also performs not well as features are less powerful therein. 

To address the above deficiencies, we make the following refactoring and develop an improved one-stage lightweight model for detection. (1) We change stride of Conv1 from $2$ to $1$. The original network design quickly downsamples the input image to reduce computational cost. But object detection requires higher resolution feature to make downstream feature decoder (the detector heads) work well. Such modification enables utilization of all convolution layers while preserves high resolution for the top feature map. (2) We modify the output channel of Conv1 from 24 to 16, which reduces memory footprint and computation. (3) We reduce the block number of stage-3 from $8$ to $6$. We find such modification leads to slightly lower pre-training precision, but does not hurt detection performance. The overall runtime is reduced  significantly. (4) We add two additional shufflenet blocks which are trained from scratch before the regression and classification head. The added blocks provide additional adaptation of the high level feature for detection. (5) We employ very simple RPN-alike detector which discriminate between classes. Unlike previous layers, the detection heads use full convolution, while parameters are increased, we find this significantly improves accuracy. We refer such lightweight base detector as $1\times$ in the following sections. Refer to the supplementary material for architecture diagram of the model.

\subsection{Imitation with lightweight detectors}
\label{exp_imitation_toy_detector}
We first apply the proposed method to the toy detector presented above. We use the base model as teacher (denoted as $1\times$), and directly halve channels of each layer for student model. Specifically, we halve once of teacher model to get the $0.5\times$  model, and halve twice (75\% channels removed) to obtain the $0.25\times$ model. We conduct the experiments on challenging KITTI~\cite{Geiger2012CVPR} dataset. Since test set annotation is not available, we follow~\cite{cai2016unified,mao2017can} to split training dataset into training and validation sets and carefully make sure they do not come from the same video sequence. We use the official evaluation tool to evaluate detector performance on the validation set.
Table~\ref{exp_toy_detector} shows overall imitation results of the student models, as well as comparison to other methods. It is well known that reduction on parameters and computation always brings exponential performance drop, \emph{e.g.}, the $0.5\times $ model sacrifices only around 4.7 mAP compared to the teacher, while $0.25\times $ halving results in $16.7$ mAP drop. In such hard cases, the presented method still achieves significant boost for student models, \emph{i.e.}, the $0.5\times$ model gets 2.5 mAP improvement, the $0.25\times $ model is boosted by 6.6 mAP ($0.25\times $-I), which is 14.7\% of the non-imitated one. Note the improvement for $0.5\times $ model on pedestrian is smaller than other classes as the gap between teacher and non-imitated student is minor on pedestrian.

We conduct experiments on 4 comparing settings with the $0.25\times $ model. As shown in last 4 lines of Table~\ref{exp_toy_detector}. The first is hint learning~\cite{romero2014fitnets} (\emph{i.e.} full feature imitation, denoted as $0.25\times $-F). Though performing well for classification, it brings large performance drop (8.9 mAP) to the original $0.25\times $ model. We conjecture this is because background  noise  overwhelms the informative supervision signal from teacher model which is verified in Sec. \ref{per_channel_variance}. The very simple setting ($0.25\times $-G) of directly scaling ground truth boxes with same stride on the feature layer and applying imitation on those areas gives much less gain than the proposed method. The reason is that while noise from background regions is avoided, the method also missed the important supervision from some near object locations. In the third setting ($0.25\times $-D), we find adapting the vanilla knowledge distillation~\cite{Hinton2015DistillingTK} to detection setting produces unpleasant result (only 0.9 increase of mAP), verifies our intuition in Sec.~\ref{intro}. Finally, we try to combine distillation loss with imitation loss (denoted as 0.25-ID), but the  performance is worse than only using imitation term, implying high level feature imitation and distillation on model outputs have very divergent objectives.

\begin{table*}[h]
	\centering
		\renewcommand{\tabcolsep}{1.3pt}
		\renewcommand{\arraystretch}{1.6}
		\small
	\begin{tabular}{l|c|cccccccccccccccccccc}
		\hline
		Model   & mAP  & aero & bike & bird & boat & bottle & bus   & car  & cat  & chair & cow  & table & dog  & horse & mbike & person & plant & sheep & sofa & train & tv   \\ \hline
		
		res101    & 74.4 & 77.8 & 78.9 & 77.5 & 63.2 & 62.6   & 79.2 & 84.4 & 85.6 & 54.5  & 81.5 & 68.7  & 85.7 & 84.6  & 77.8  & 78.6   & 47.1  & 76.3  & 74.9 & 78.8  & 71.2 \\ 
		res101h  & 67.4 & 73.9& 78.6& 66.3& 52.5& 42.4& 73.8& 80.4& 80.1& 43.5& 71.8& 61.9& 78.7& 81.7& 74.4& 76.8& 42.2& 66.9& 65. & 74.3& 62.8 \\ 
		res101h-I & 71.2 & 77.2& 80.0 & 72.9& 56.0 & 50.4& 77.1& 82.3& 85.5& 47.4& 80.2& 59.9& 84.3& 83.9& 73.8& 79.1& 44.6& 70.8& 69.4& 78.7& 70.4 \\ 
		&\textbf{ +3.8} & \textbf{+3.3} & \textbf{+ 1.4} & \textbf{+ 6.6} & \textbf{+ 3.5} & \textbf{+ 8.0} & \textbf{+ 3.3} & \textbf{+ 1.9} & \textbf{+ 5.4} & \textbf{+ 3.9}  & \textbf{+ 8.4} & -2.0  & \textbf{+ 5.6} & \textbf{+ 2.2}  &  -0.6  & \textbf{+ 2.3}  & \textbf{+ 2.4}  & \textbf{+ 3.9}  & \textbf{+ 4.4} & \textbf{+ 4.4}  & \textbf{+ 7.6} \\ \hline
	\end{tabular}
	\caption{Imitation with halved student model with Faster R-CNN model on Pascal VOC07 dataset.}
	\label{exp_halved_student_pascal}
	\vspace{-2mm}
\end{table*}

\begin{table*}[h]
	\centering
	\renewcommand{\tabcolsep}{1.3pt}
	\renewcommand{\arraystretch}{1.6}
	\small 
	\begin{tabular}{l|c|cccccccccccccccccccc}
		\hline
		Model   & mAP  & aero & bike & bird & boat & bottle & bus   & car  & cat  & chair & cow  & table & dog  & horse & mbike & person & plant & sheep & sofa & train & tv   \\ \hline
		VGG16   & 70.4 & 70.9 & 78.0 & 67.8 & 55.1 & 53.2   & 79.6  & 85.5 & 83.7 & 48.7  & 78.0 & 63.5  & 80.2 & 82.0  & 74.5  & 77.2   & 43.0  & 73.7  & 65.8 & 76.0  & 72.5 \\ 
		VGG11   & 59.6 & 67.3 & 71.4 & 56.6 & 44.3 & 39.3   & 68.8  & 78.4 & 66.6 & 37.7  & 63.2 & 51.6  & 58.3 & 76.4  & 70.0  & 71.9   & 32.2  & 58.1  & 57.8 & 62.9  & 60.0 \\
		VGG11-I & 67.6 & 72.5 & 73.8 & 62.8 & 53.1 & 49.2   & 80.5  & 82.7 & 76.8 & 44.8  & 73.5 & 64.3  & 72.6 & 81.1  & 75.3  & 76.3   & 40.2  & 66.3  & 61.8 & 73.4  & 70.6 \\ 
		& \textbf{+8.0} & \textbf{+5.2}&  \textbf{+2.4}&  \textbf{+6.2}&  \textbf{+8.8}&  \textbf{+9.9}& \textbf{+11.7}&  \textbf{+4.3}& \textbf{+10.2}&  \textbf{+7.1}& \textbf{+10.3}& \textbf{+12.7}& \textbf{+14.3}&  \textbf{+4.7}&  \textbf{+5.3}&  \textbf{+4.4}&  \textbf{+8.0} &  \textbf{+8.2}&  \textbf{+4.0} & \textbf{+10.5}& \textbf{+10.6}  \\ \hline

		res101    & 74.4 & 77.8 & 78.9 & 77.5 & 63.2 & 62.6   & 79.2 & 84.4 & 85.6 & 54.5  & 81.5 & 68.7  & 85.7 & 84.6  & 77.8  & 78.6   & 47.1  & 76.3  & 74.9 & 78.8  & 71.2 \\ 
		res50   & 69.1 & 68.9& 79.0 & 67.0 & 54.1& 51.2& 78.6& 84.5& 81.7& 49.7& 74.0 & 62.6& 77.2& 80. & 72.5& 77.2& 40.0 & 71.7& 65.5& 75.0 & 71.0 \\ 
		res50-I & 72.0 & 71.5& 80.6& 71.1& 57.0 & 52.4& 82.1& 90.0 & 82.7& 51.6& 74.5& 66.2& 82.3& 82.3& 75.7& 78.3& 43.5& 79.6& 69.1& 77.3& 72.1 \\ 
		&\textbf{ +2.9} & \textbf{+2.6} & \textbf{+1.6} & \textbf{+4.1} & \textbf{+2.9} & \textbf{+1.2}   & \textbf{+3.5} & \textbf{+5.0} & \textbf{+1.0} & \textbf{+1.9}  & \textbf{+0.5} & \textbf{+3.6}  & \textbf{+5.1} & \textbf{+2.3}  & \textbf{+3.2}  & \textbf{+1.1}   & \textbf{+3.5}  & \textbf{+7.9}  & \textbf{+3.6} & \textbf{+2.3}  & \textbf{+1.1} \\ \hline
	\end{tabular}
	\caption{Imitation with shallow student model on Pascal-VOC07 dataset with Faster R-CNN model.}
	
	\label{shallow_student}
	\vspace{-2mm}
\end{table*}

\begin{table}
	\small
	\renewcommand{\tabcolsep}{1.5pt}
	\renewcommand{\arraystretch}{1.1}
	\begin{tabular}{l|c|c|ccc|c|ccc}
		\hline
		Model     & AP@0.5 & AP   & AP$_{s}$  & AP$_{m}$  & AP$_{l}$  & AR   & AR$_{s}$  & AR$_{m}$  & AR$_{l}$  \\ \hline
		res101    & 54.6   & 34.4 & 14.3 & 39.1 & 51.9 & 45.9 & 23.0 & 52.2 & 66.4 \\ 
		res101h   & 48.4   & 28.8 & 11.8 & 32.0 & 44.9 & 41.5 & 19.8 & 45.9 & 62.3 \\ 
		res101h-I & 51.2& 31.6& 13.2& 35.9& 47.5& 44.0 & 22.4& 50.3& 64.5 \\ 
		&\textbf{+2.8} & \textbf{+2.8} &  \textbf{+1.4}&  \textbf{+3.9}& \textbf{ +2.6}&  \textbf{+2.5}&  \textbf{+2.6}&  \textbf{+4.4}&  \textbf{+2.2} \\  \hline
	\end{tabular}
	\caption{Imitation with halved student model with Faster R-CNN model on COCO dataset.}
	\label{exp_halved_student_COCO}
	\vspace{-4mm}
\end{table}

\begin{table}[h]
	\centering
	\small
	\renewcommand{\tabcolsep}{2.0pt}
	\renewcommand{\arraystretch}{1.1}
	\begin{tabular}{l|c|c|ccc|c|ccc}
		\hline
		Model     & AP@0.5 & AP   & AP$_{s}$  & AP$_{m}$  & AP$_{l}$  & AR   & AR$_{s}$  & AR$_{m}$  & AR$_{l}$  \\ \hline
		res50 & 59.0 & 36.9 & 21.5 & 39.8 & 48.3 & 50.5 & 31.4 & 53.9 & 63.6 \\ 
		res50h   & 52.6& 31.2& 18.5& 32.0& 42.4& 46.3& 27.7& 47.5& 60.6 \\ 
		res50h-I & 55.8 & 34.8 & 21.0 & 34.9 & 45.5 & 49.1 & 30.5 & 52.6 & 63.5 \\ 
		&\textbf{+3.2}& \textbf{+3.6}& \textbf{+2.5}& \textbf{+2.9}& \textbf{+3.1}& \textbf{+2.8}& \textbf{+2.8}& \textbf{+5.1}& \textbf{+2.9} \\  \hline	
	\end{tabular}
	\caption{Result of multi-layer imitation on COCO dataset with Resnet50 FPN based Faster R-CNN model.}	
	\label{exp_mutli_layer_imitation}
	\vspace{-4mm}
\end{table}

\subsection{Imitation with Faster R-CNN}
We further perform extensive experiments with the more general architecture of Faster R-CNN model under three settings: 1) halved student model. 2) shallow student model. 3) multi-layer imitation.
\vspace{-3mm}
\paragraph{Halved student model}
In this setting, we use Resnet101 based Faster R-CNN as teacher model and halve channel number of each layer including the fully connected layers to construct the student model. As shown in Table~\ref{exp_halved_student_COCO} and Table~\ref{exp_halved_student_pascal}, we perform experiments with COCO and Pascal VOC07 dataset. Clearly halving the whole teacher model cause the performance to drop significantly. With imitation, the halved student model gets significant boost, \emph{i.e.}, 2.8 absolute mAP gain both in Pascal style average precision and COCO style average precision with COCO dataset; and 3.8 absolute mAP gain for Pascal VOC07 dataset. The results demonstrate that our method can effectively distill the teacher detector's knowledge into the halved student.

\vspace{-3mm}
\paragraph{Shallow student network}
For this setting, instead of halving layer channels of teacher model, we choose shallower student backbone with similar architecture of teacher model. Specifically, we perform two imitation experiments: VGG11 based Faster R-CNN as student and VGG16 based one as teacher;  Resnet50 based Faster R-CNN as student and Resnet101 based one as teacher. As shown in Table~\ref{shallow_student}, the shallow backbone based student model all gets significant improvement, especially for the VGG11 based student model, the imitated model gets 8.0 absolute gain in mAP, our method nearly recovers \textbf{74\%} of the performance drop due to shallow backbone.

\vspace{-3mm}
\paragraph{Multi-layer imitation}
The previous imitation experiments are with single layer of feature map, we further extend the experiment to multi-layer imitation with seminal work of Feature Pyramid Networks (FPN)~\cite{lin2017feature}. The FPN combined with Faster R-CNN framework perform region proposal on different layer with different anchor prior size, and pools feature on corresponding layer according to roi size. We compute the imitation region on each layer with corresponding prior anchors, and let student model imitate feature response on each layer. The teacher detection model is a Resnet50 FPN based Faster R-CNN, and student is a halved counterpart. As shown in Table~\ref{exp_mutli_layer_imitation}, imitated student gets 3.2 absolute mAP gain in Pascal style average precision and 3.6 mAP gain with COCO style average precision.

\subsection{Analysis}

\subsubsection{Visualization of imitation mask}
\begin{figure*}[!t]
	\centering
	\subfigure[]{ 
		\label{fig_original_image} 
		\begin{minipage}[b]{0.32\textwidth} 
			\centering 
			\includegraphics[width=1\linewidth]{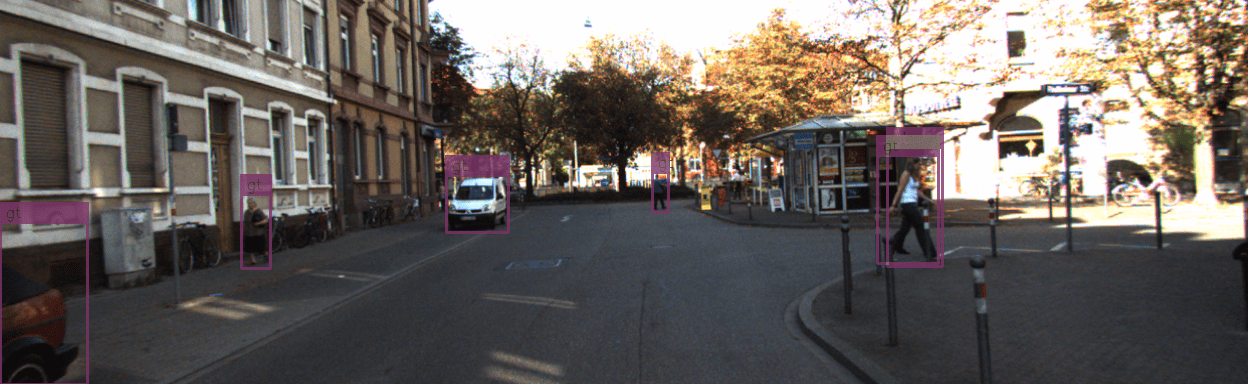} 
	\end{minipage}}%
	\subfigure[]{ 
		\label{fig_t0.2} 
		\begin{minipage}[b]{0.32\textwidth} 
			\centering 
			\includegraphics[width=1\linewidth, ]{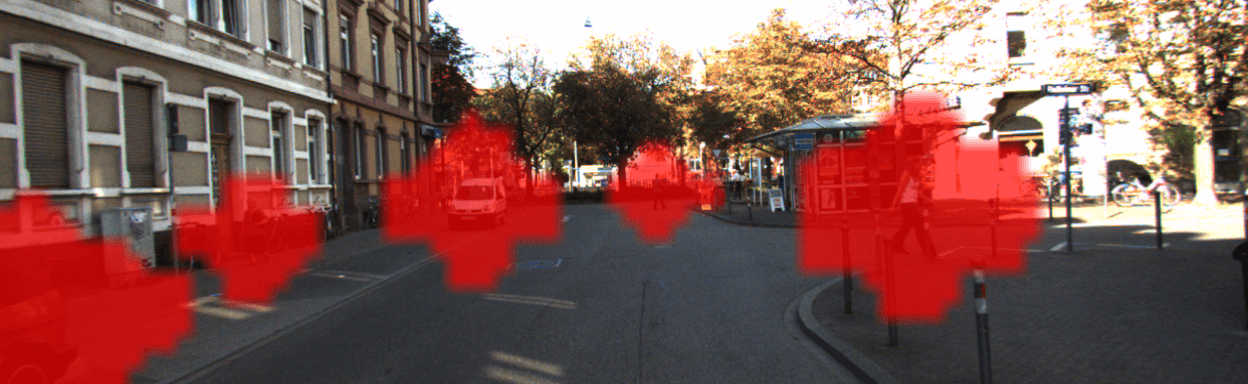} 
	\end{minipage}} 
	\subfigure[]{ 
		\label{fig_t0.5} 
		\begin{minipage}[b]{0.32\textwidth} 
			\centering 
			\includegraphics[width=1\linewidth, ]{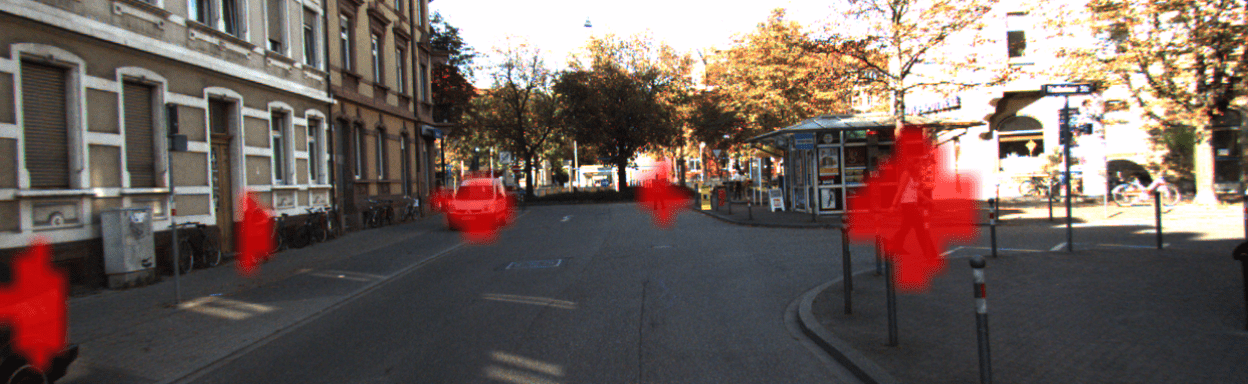} 
	\end{minipage}}
	\subfigure[]{ 
		\label{fig_t0.8} 
		\begin{minipage}[b]{0.32\textwidth} 
			\centering 
			\includegraphics[width=1\linewidth, ]{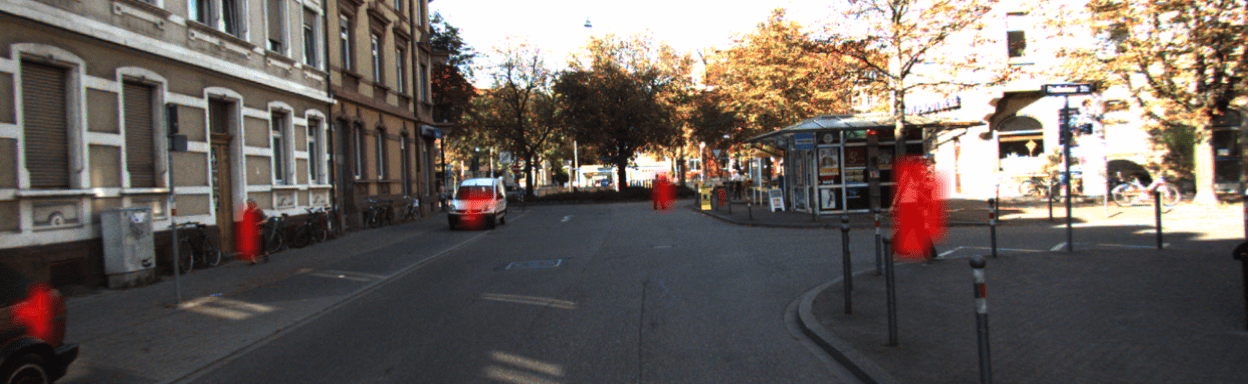} 
	\end{minipage}} 
	\subfigure[]{ 
		\label{fig_ht0.5} 
		\begin{minipage}[b]{0.32\textwidth} 
			\centering 
			\includegraphics[width=1\linewidth, ]{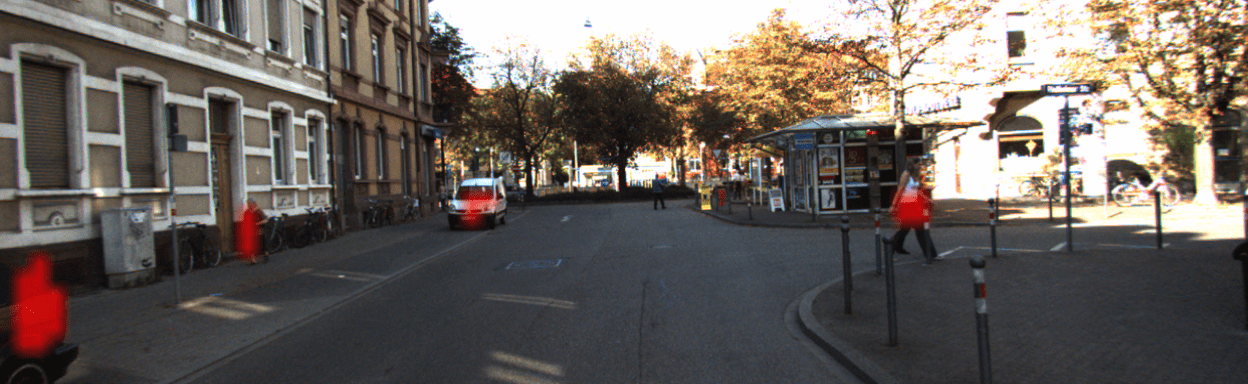} 
	\end{minipage}} 
	\subfigure[]{ 
		\label{fig_ht0.8} 
		\begin{minipage}[b]{0.32\textwidth} 
			\centering 
			\includegraphics[width=1\linewidth, ]{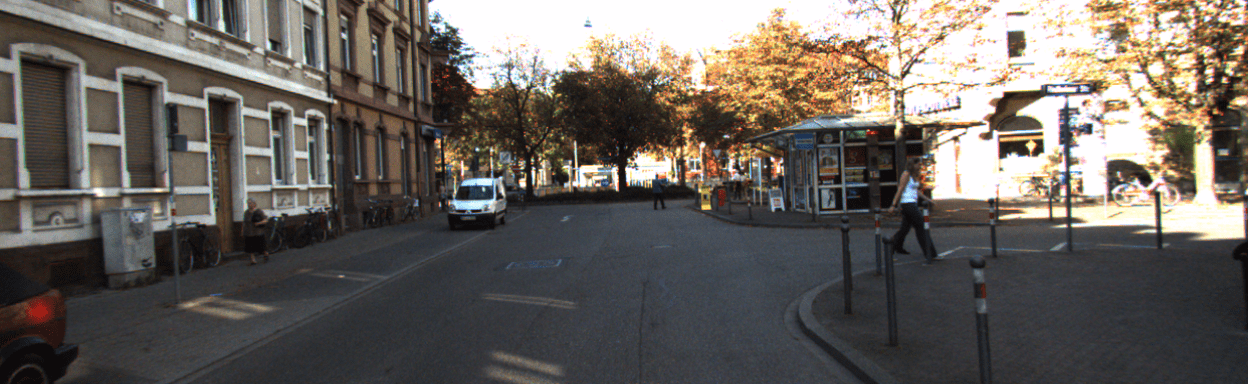} 
	\end{minipage}} 			
	\caption{Examples of calculated imitation masks overlaid on input image. Note that the actual masks are calculated on last feature map, we enlarge the mask with corresponding ratio to display on the input image. (a) Original image. (b) $\psi=0.2$. (c) $\psi=0.5$. (d) $\psi=0.8$. (e) Hard-thresh-0.5. (f) Hard-thresh-0.8. Thresh-* indicates different thresholding factor for proposed approach, Hard-thresh-* means using constant threshold of $F$ when filtering the IOU map.} 
	\label{fig_sample_imitation_mask} 
	\vspace{-4mm}
\end{figure*}

To better understand the imitation region generated by our approach,  we visualize some example masks $ I $ on input image with the toy detector given sample from KITTI dataset. Specifically we scale the generated imitation mask $ I $ on the feature map to input image with corresponding stride(16 for the toy detector). Fig~\ref{fig_sample_imitation_mask} shows example imitation masks scaled and overlaid on input image. Of the 6 images, Fig~\ref{fig_original_image} is original image; Fig~\ref{fig_t0.2}~\ref{fig_t0.5}~\ref{fig_t0.8} are generated with $ \psi=0.2 $, $ \psi=0.5 $, and $ \psi=0.8 $ respectively; Fig~\ref{fig_ht0.5}~\ref{fig_ht0.8} are filtered with constant threshold value of $ F=0.5 $ and $ F=0.8 $ respectively. It is obvious that some objects are missing with only $ F=0.5 $, and nearly all imitation mask disappeared with $ F=0.8 $. This is because constant filter threshold of $ F $ biases for those ground truth boxes of similar size with prior anchors. Our method with adaptive filter threshold greatly mitigates this problem.

\vspace{-3mm}
\subsubsection{Qualitative performance gain from imitation}\label{exp_imitation_gain_qualitative}
\begin{figure*}[!t]
\centering
\subfigure[]{ 
\label{qualitative_discrimination_student} 
\begin{minipage}[b]{0.19\textwidth} 
\centering 
\includegraphics[width=1.\linewidth,height=1.4in]{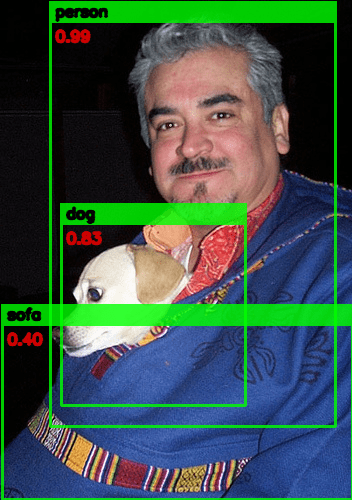}
\end{minipage}} 
\subfigure[]{ 
\label{qualitative_localization_student} 
\begin{minipage}[b]{0.19\textwidth} 
	\centering 
	\includegraphics[width=1.\linewidth,height=1.4in]{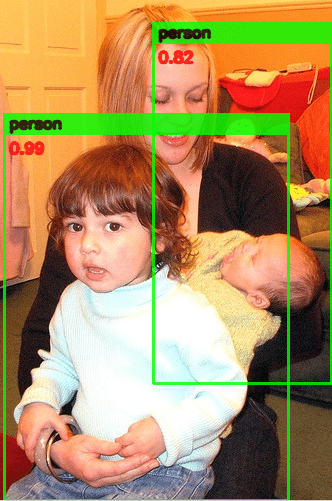}
\end{minipage}} 
\subfigure[]{ 
	\label{qualitative_repeated_detection_student} 
	\begin{minipage}[b]{0.19\textwidth} 
		\centering 
		\includegraphics[width=1.\linewidth,height=1.4in]{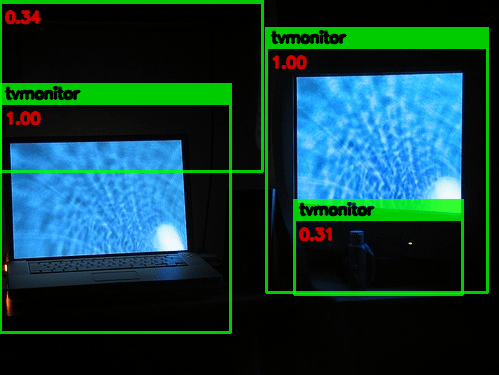}
	\end{minipage}} 
	\subfigure[]{ 
		\label{qualitative_background_error_student} 
		\begin{minipage}[b]{0.19\textwidth} 
			\centering 
			\includegraphics[width=1.\linewidth,height=1.4in]{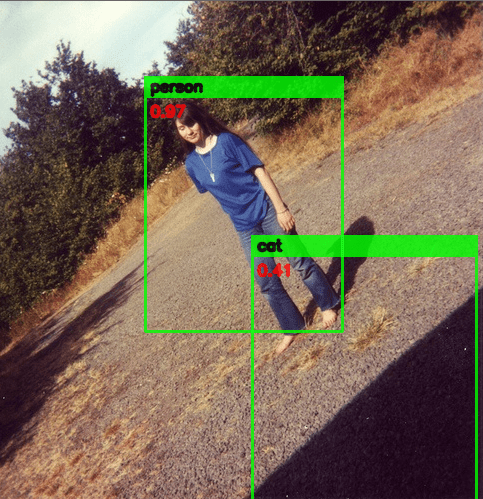}
		\end{minipage}} 
		\subfigure[]{ 
			\label{qualitative_grouped_detection_student} 
			\begin{minipage}[b]{0.19\textwidth} 
				\centering 
				\includegraphics[width=1.\linewidth,height=1.4in]{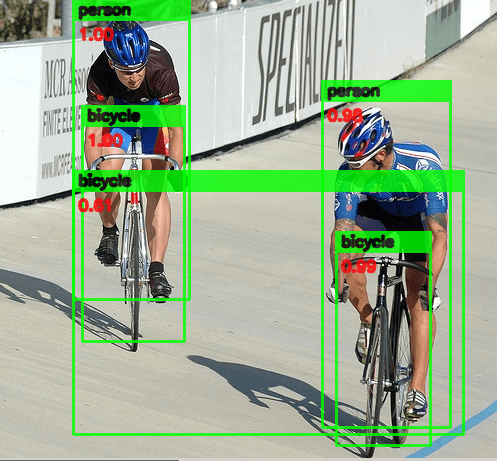}
			\end{minipage}}
			
			\subfigure[]{ 
				\label{qualitative_discrimination_teacher} 
				\begin{minipage}[b]{0.19\textwidth} 
					\centering 
					\includegraphics[width=1.\linewidth,height=1.4in]{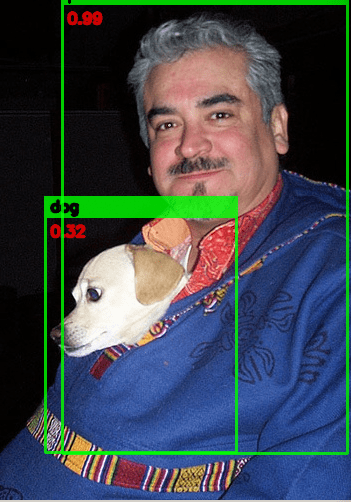}
				\end{minipage}} 
				\subfigure[]{ 
					\label{qualitative_localization_teacher} 
					\begin{minipage}[b]{0.19\textwidth} 
						\centering 
						\includegraphics[width=1.\linewidth,height=1.4in]{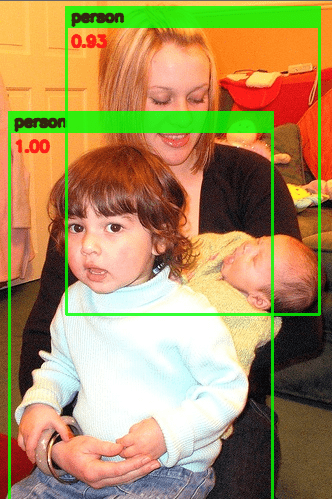}
					\end{minipage}} 
					\subfigure[]{ 
						\label{qualitative_repeated_detection_teacher} 
						\begin{minipage}[b]{0.19\textwidth} 
							\centering 
							\includegraphics[width=1.\linewidth,height=1.4in]{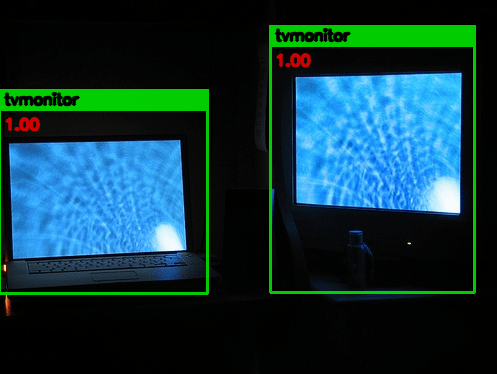}
						\end{minipage}} 
						\subfigure[]{ 
							\label{qualitative_background_error_teacher} 
							\begin{minipage}[b]{0.19\textwidth} 
								\centering 
								\includegraphics[width=1.\linewidth,height=1.4in]{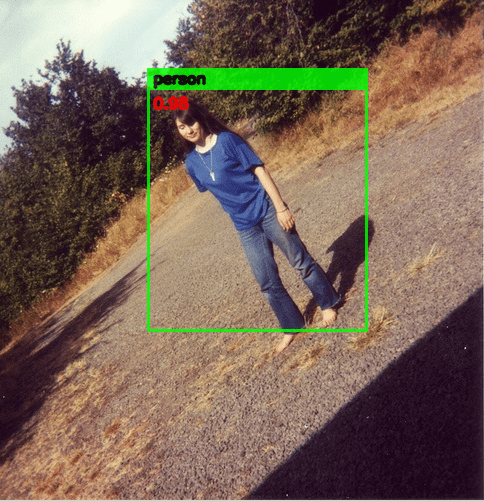}
							\end{minipage}} 
							\subfigure[]{ 
								\label{qualitative_grouped_detection_teacher} 
								\begin{minipage}[b]{0.19\textwidth} 
									\centering 
									\includegraphics[width=1.\linewidth,height=1.4in]{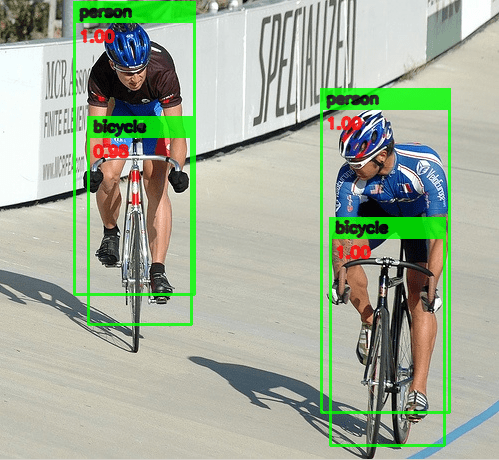}
								\end{minipage}} 			
								\caption{Qualitative results on the gain from imitation learning. The bounding box visualization threshold is set as 0.3. The top row images are student model's output without imitation, the bottom row shows imitated student's output. }
								\label{qualitative_result}
								\vspace{-3mm}
							\end{figure*}
			\begin{figure*}[!t]
				\subfigure[]{ 
					\label{pie_furniture} 
					\begin{minipage}{0.32\textwidth} 
						\centering 
						\includegraphics[width=0.49\linewidth]{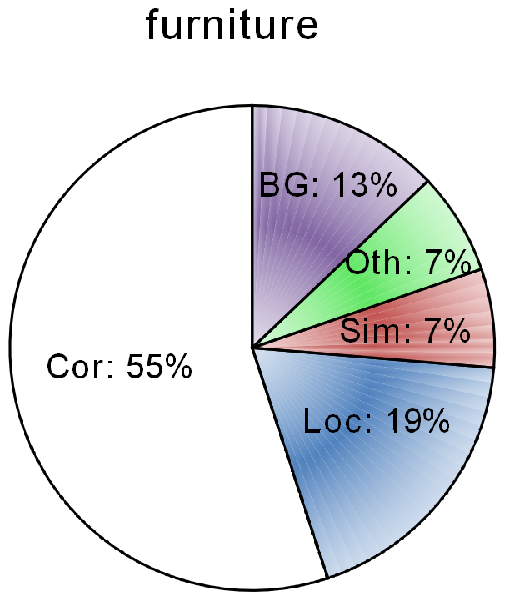} 
						\hspace{-10pt}
						\includegraphics[width=0.49\linewidth]{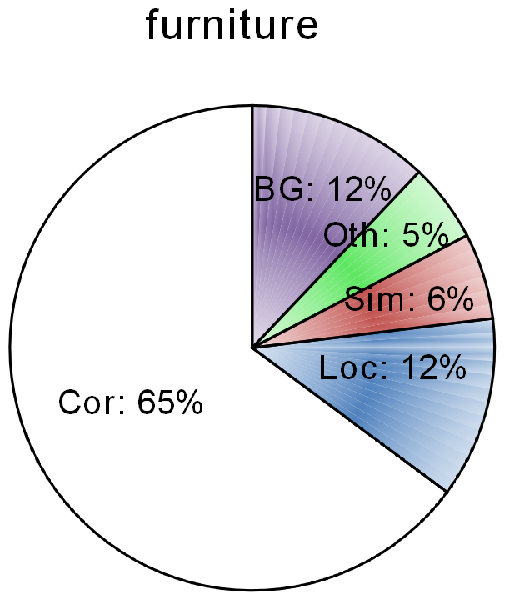} 
					\end{minipage}}%
					\subfigure[]{ 
						\label{pie_animals} 
						\begin{minipage}{0.32\textwidth} 
							\centering 
							\includegraphics[width=0.49\linewidth]{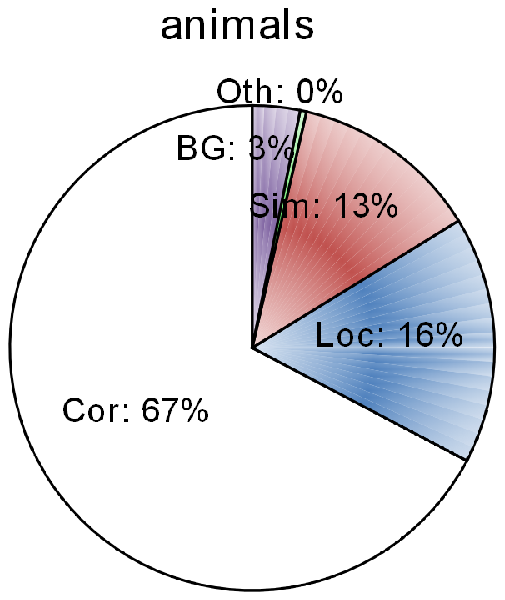} 
							\hspace{-10pt}
							\includegraphics[width=0.49\linewidth]{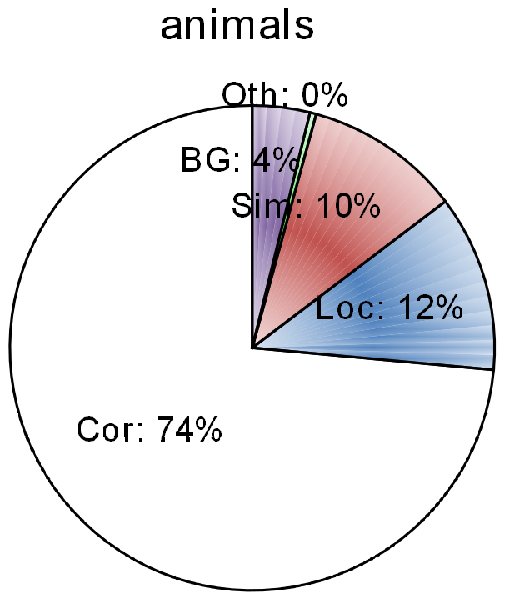} 
						\end{minipage}}%
						\subfigure[]{ 
							\label{pie_vehicles} 
							\begin{minipage}{0.32\textwidth} 
								\centering 
								\includegraphics[width=0.49\linewidth]{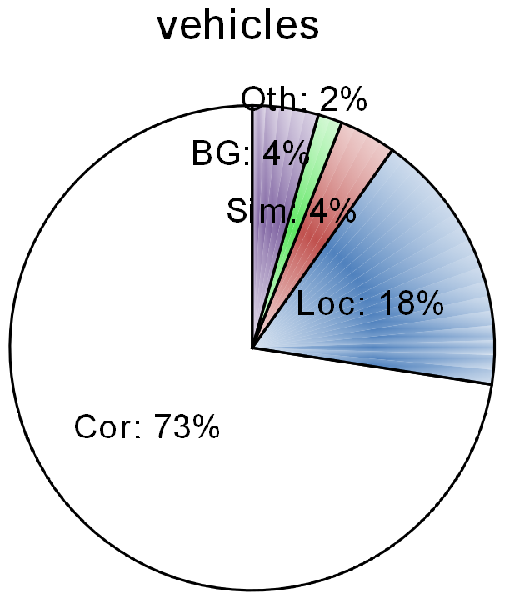} 
								\hspace{-10pt}
								\includegraphics[width=0.49\linewidth]{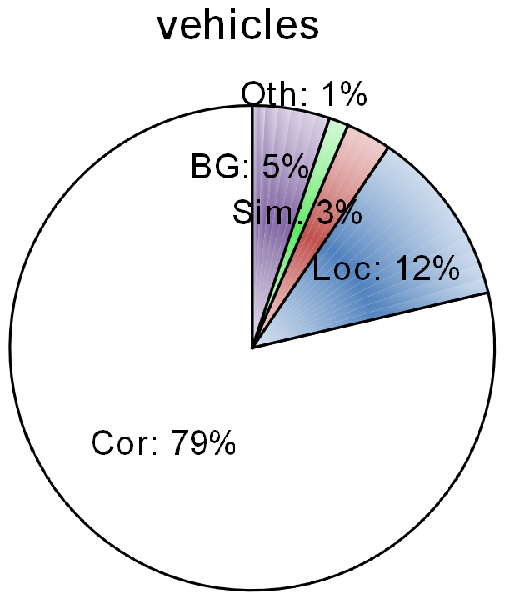} 
							\end{minipage}}%
							\caption{Imitation gain from error perspective with VGG11 based Faster R-CNN student and VGG16 based teacher on the Pascal VOC07 dataset. For each pair, the left figure corresponds to raw student model, and the right corresponds to imitated student.}
							\label{error_analysis_imitated_non_imitated} 
							\vspace{-3mm}
						\end{figure*}
In this subsection, we present some sampled detection outputs reflecting the enhanced ability of student detector through the imitation learning. The results are from VGG11 based Faster R-CNN model on VOC07 dataset (ref. to Table~\ref{shallow_student} for quantitative results). We only show one example for each type of gain due to space limited, and choose the examples containing simple objects for clearer visualization. In Fig~\ref{qualitative_result}, the upper row of detection outputs are from raw student model trained with ground truth supervision only, and the lower row of detection outputs are from imitated student model. The improvement of the student model with teacher supervision can be summarized into following aspects:
\emph{Improved discrimination ability.} As shown in Fig~\ref{qualitative_discrimination_student} and Fig~\ref{qualitative_discrimination_teacher}, the color and style of lower part of the man's clothes is somewhat similar to that in some sofa objects. The raw student model mistakingly detect that as a sofa object with rather high confidence. While the imitated student avoids the error, indicating better discrimination ability. It is interesting to note that the imitated student has lower confidence on the dog instance compared to the raw student model, we have observed the teacher model (VGG16 based Faster R-CNN) outputs confidence of 0.38 for the instance. This phenomenon reveals that the teacher model's learned knowledge has been effectively transfered to the student model.
\emph{More reliable localization.} As shown in Fig~\ref{qualitative_localization_student} and Fig~\ref{qualitative_localization_teacher}, the raw student model outputs a rather inaccurate location of the woman as a person instance. While the imitated student model learns better localization knowledge from the teacher and outputs a rather accurate bounding box for the person instance.
\emph{Less repeated detection.} As shown in Fig~\ref{qualitative_repeated_detection_student} and Fig~\ref{qualitative_repeated_detection_teacher}, the raw student model outputs repeated detections for the tv-monitors which are unfortunately not able to be suppressed by NMS. While the imitated model predicts single bounding box for each object. This phenomenon indicates imitated student has better ability handling close to object input regions, this improvement comes from improved region proposal and enhanced ROI processing ability.
\emph{Less background error.} As shown in Fig~\ref{qualitative_background_error_student} and Fig~\ref{qualitative_background_error_teacher}, the raw student model wrongly predict an area of background as a cat instance. While the imitated the student avoids the error, indicating lower background false positive prediction.
\emph{Avoiding grouped detection error.} We have observed grouped detection of near objects is a common error case for the raw student model, as shown in left image of Fig~\ref{qualitative_grouped_detection_student} and Fig~\ref{qualitative_grouped_detection_teacher}. The imitated student gets improved ability in avoiding such error case.

			\begin{figure*}[!t]
				\centering
				\subfigure[]{ 
					\label{fig_thresholds} 
					\begin{minipage}[b]{0.32\textwidth} 
						\centering 
						\includegraphics[width=1\linewidth]{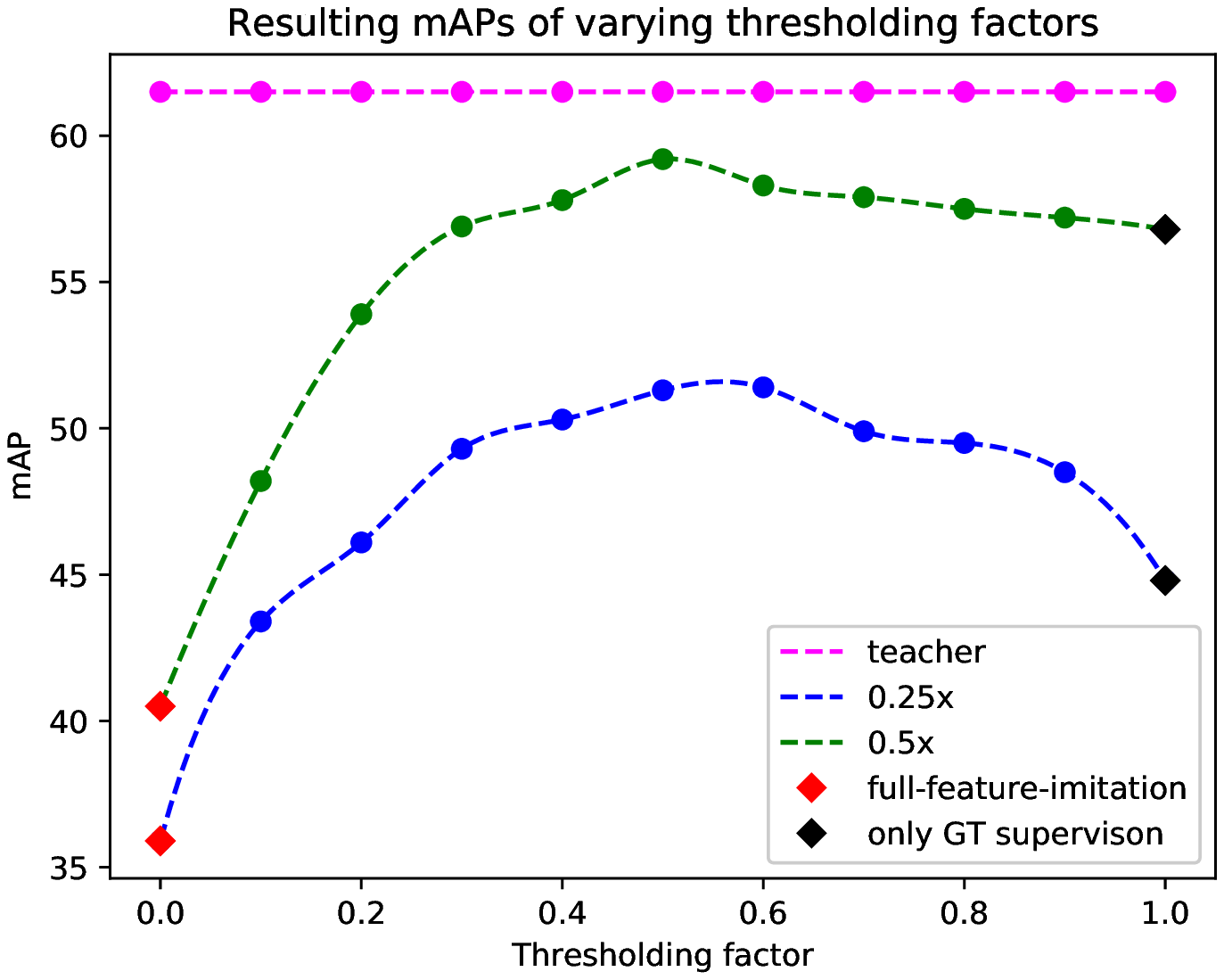} 
				\end{minipage}}%
				\subfigure[]{ 
					\label{fig_variance_toy_model_kitti} 
					\begin{minipage}[b]{0.32\textwidth} 
						\centering 
						\includegraphics[width=1\linewidth]{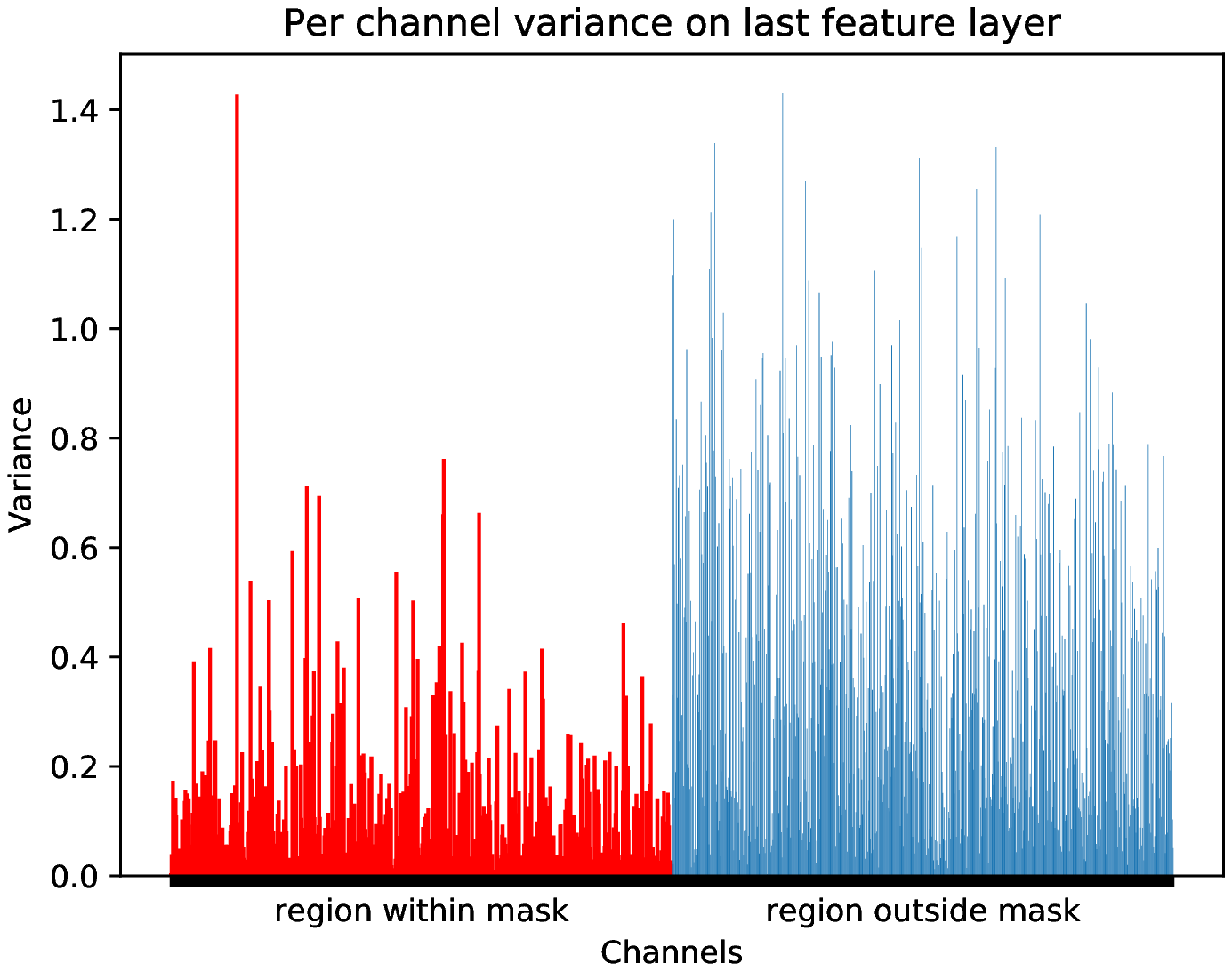} 
				\end{minipage}} 
				\subfigure[]{ 
					\label{fig_variance_frcnn_coco} 
					\begin{minipage}[b]{0.32\textwidth} 
						\centering 
						\includegraphics[width=1\linewidth]{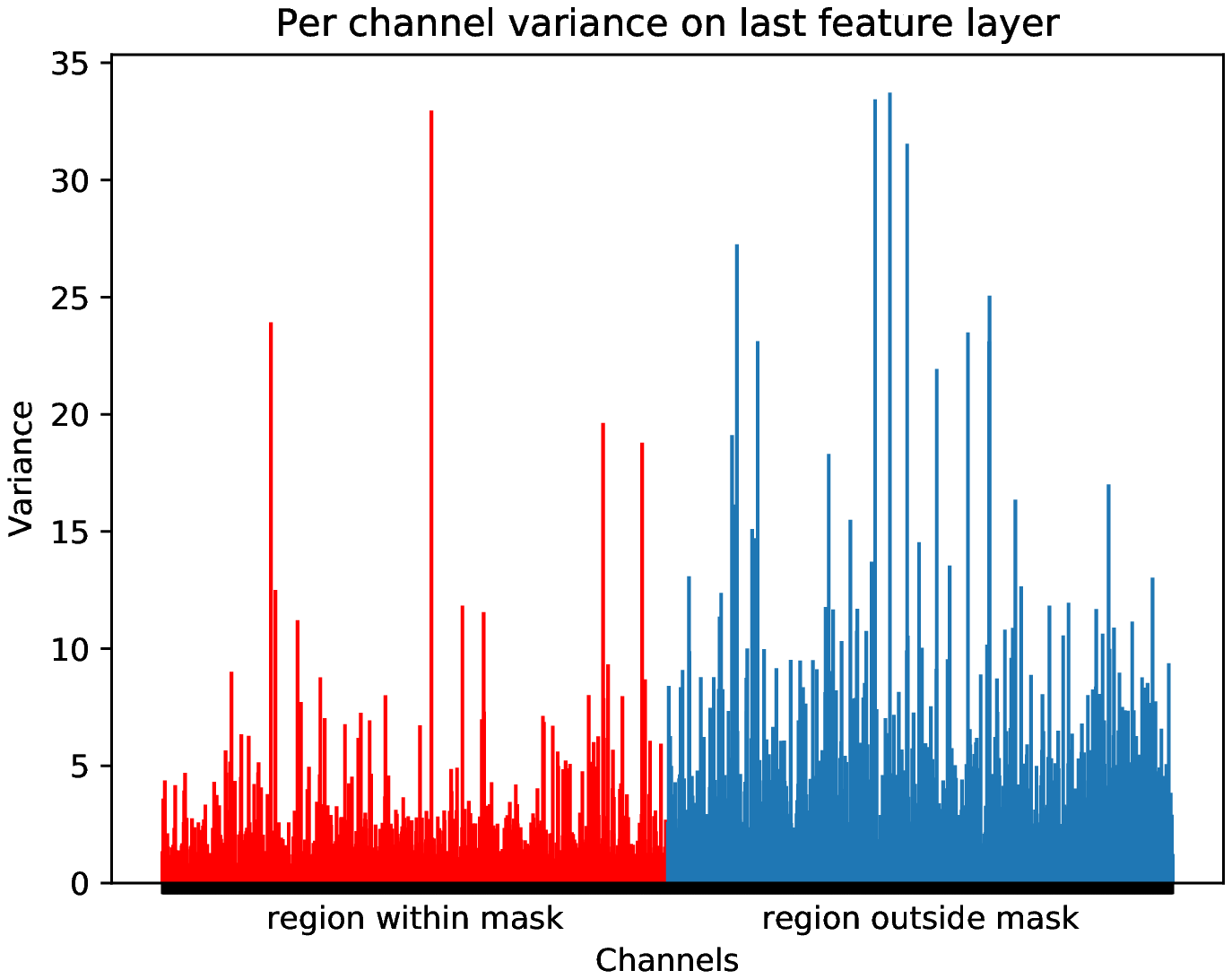} 
				\end{minipage}} 
				\caption{Results for further investigation of the method. (a) Varying imitation thresholding factor $ \psi $ for the toy detector experiment. (b),(c) Per-channel variance on high level feature map of learned teacher model. (b) is calculated with toy detector on KITTI dataset.(c) is calculated with Faster R-CNN on COCO dataset.}
				\label{fig_further_investigation} 
				\vspace{-4mm}
			\end{figure*}		

\vspace{-3mm}
\subsubsection{Quantitative performance gain from imitation}\label{exp_imitation_gain_error_perspective}

We use the analysis tool from ~\cite{hoiem2012diagnosing} to understand the type of detection errors reduced by imitating teacher model. The analysis is performed with the VGG11 Faster R-CNN student on Pascal VOC07 dataset (the teacher model is VGG16 based Faster R-CNN, ref. to Table~\ref{shallow_student} for average precision gain results). We present analysis on 3 grouped object class set: 1) vehicles. 2)  animals containing all animals including person. 3) furniture including chair, dining table and sofa. The detections were classified into five groups: 
\textbf{1)} Correct detection(Cor): correct class and $ IoU > 0.5 $.
\textbf{2)} Localization (Loc): correct class, but misaligned bounding box ($ 0.1 < IoU < 0.5 $). 
\textbf{3)} Similar (Sim): wrong class, correct category, $IoU > 0.1$.
\textbf{4)} Other (Oth): wrong class and category, $ IoU > 0.1 $. 
\textbf{5)} Background (BG): $ IoU < 0.1 $ for any object class.
Due to limited space we only present pie chart error percentage result, and defer to supplementary file for other analysis result. 
As shown in Fig~\ref{error_analysis_imitated_non_imitated}, we observe for the three sub-set of object class, our method significantly improves the number of correct detections, and effectively reduces all other kinds of detection errors, especially for the Loc term. The error  composition analysis reveals following important improvements: 1) Stronger localization ability (Loc); 2) less confusion between the same category and other category objects (Sim and Oth); 3) less background induced errors (BG).

\vspace{-3mm}
\subsubsection{Varying $ \psi $ for generating mask}
\label{vary_psi} 
To investigate the effects of region selection for imitation, we perform experiments on the $0.5\times$ and $0.25\times$ student models with varying thresholding factor $ \psi $. We record mean value among three runs and plot the performance curve in Fig.~\ref{fig_thresholds}
When $ \psi =0$ , all points will be preserved and the method degenerates to full feature imitation as in hint learning. It is clear that imitated models are misguided severely. The mAP is even much lower than the ones trained with only ground truth supervision. As the threshold value increases, the student model performs much better, even with very low threshold of $0.1$. This is strong evidence that the proposed approach effectively finds useful information while filters detrimental knowledges. The neutral value of $0.5$ turns out to be optimal. 
When $ \psi $ is larger than $0.5$, both students' mAP starts decreasing, but all the way still higher than when the value is $1.0$, under which case the imitation reduces to only ground truth supervision. It is worth noting that when the $ \psi $ is larger than 0.5, the imitation regions quickly shrink and become extremely tiny and sparse, but imitation on those area still significantly boosts the students.

\vspace{-3mm}
\subsubsection{Per-channel variance of high level responses}
\label{per_channel_variance} 
To understand why full feature imitation produces deteriorated performance, we calculate the per-channel variance of the imitation feature map from a trained teacher model. We randomly sample and pass 10 images through the teacher model, calculate and record variances for anchor location within imitation region (with $\psi = 0.5$) and outside the region for each channel separately. Results are shown in Fig~\ref{fig_variance_toy_model_kitti} and Fig~\ref{fig_variance_frcnn_coco} for the KITTI and COCO dataset on our $1\times $ toy detector and Resnet101 based Faster R-CNN model. Clearly the variances under the regions selected with proposed approach are smaller than those outside the areas, and holds for nearly all channels. This indicates that responses on background areas contain much noise. Features from the regions within the mask  are  more informative. Since convolution shares weights for whole feature map, directly imitating global feature responses would unavoidably accumulate large amount of noisy gradients from background areas.  We also empirically observed that the loss value of full feature imitation is more than ten times that of proposed approach throughout training with same normalization method, which corroborates the analysis. 

\section{Conclusion}
In this work, we developed a simple to implement \emph{fine-grained feature imitation} method which employs the \emph{inter-location} discrepancy of teacher detection model's feature response on near object anchor locations to distill the knowledge in a cumbersome object detector into a smaller one. Extensive experiments and analysis demonstrate the effectiveness of our method. Importantly, the method is orthogonal to and can be further combined with other model acceleration method including pruning and quantization.
\vspace{-10pt}
\paragraph{Acknowledgement}							
Jiashi Feng was partially supported by NUS IDS R-263-000-C67-646,  ECRA R-263-000-C87-133 and MOE Tier-II R-263-000-D17-112.					
{\small
\bibliographystyle{ieee_fullname}
\bibliography{egbib}

\begin{thebibliography}{10}\itemsep=-1pt

\bibitem{alvarez2016learning}
Jose~M Alvarez and Mathieu Salzmann.
\newblock Learning the number of neurons in deep networks.
\newblock In {\em Advances in Neural Information Processing Systems}, pages
  2270--2278, 2016.

\bibitem{anisimov2017towards}
Dmitriy Anisimov and Tatiana Khanova.
\newblock Towards lightweight convolutional neural networks for object
  detection.
\newblock In {\em Advanced Video and Signal Based Surveillance (AVSS), 2017
  14th IEEE International Conference on}, pages 1--8. IEEE, 2017.

\bibitem{cai2016unified}
Zhaowei Cai, Quanfu Fan, Rogerio~S Feris, and Nuno Vasconcelos.
\newblock A unified multi-scale deep convolutional neural network for fast
  object detection.
\newblock In {\em European Conference on Computer Vision}, pages 354--370.
  Springer, 2016.

\bibitem{cai2017cascade}
Zhaowei Cai and Nuno Vasconcelos.
\newblock Cascade r-cnn: Delving into high quality object detection.
\newblock {\em arXiv preprint arXiv:1712.00726}, 2017.

\bibitem{chen2017learning}
Guobin Chen, Wongun Choi, Xiang Yu, Tony Han, and Manmohan Chandraker.
\newblock Learning efficient object detection models with knowledge
  distillation.
\newblock In {\em Advances in Neural Information Processing Systems}, pages
  742--751, 2017.

\bibitem{chen2017darkrank}
Yuntao Chen, Naiyan Wang, and Zhaoxiang Zhang.
\newblock Darkrank: Accelerating deep metric learning via cross sample
  similarities transfer.
\newblock {\em arXiv preprint arXiv:1707.01220}, 2017.

\bibitem{everingham2010pascal}
Mark Everingham, Luc Van~Gool, Christopher~KI Williams, John Winn, and Andrew
  Zisserman.
\newblock The pascal visual object classes (voc) challenge.
\newblock {\em International journal of computer vision}, 88(2):303--338, 2010.

\bibitem{Geiger2012CVPR}
Andreas Geiger, Philip Lenz, and Raquel Urtasun.
\newblock Are we ready for autonomous driving? the kitti vision benchmark
  suite.
\newblock In {\em Conference on Computer Vision and Pattern Recognition
  (CVPR)}, 2012.

\bibitem{girshick2015fast}
Ross Girshick.
\newblock Fast r-cnn.
\newblock In {\em Proceedings of the IEEE international conference on computer
  vision}, pages 1440--1448, 2015.

\bibitem{girshick2014rich}
Ross Girshick, Jeff Donahue, Trevor Darrell, and Jitendra Malik.
\newblock Rich feature hierarchies for accurate object detection and semantic
  segmentation.
\newblock In {\em Proceedings of the IEEE conference on computer vision and
  pattern recognition}, pages 580--587, 2014.

\bibitem{Girshick2015FastR}
Ross~B. Girshick.
\newblock Fast r-cnn.
\newblock {\em 2015 IEEE International Conference on Computer Vision (ICCV)},
  pages 1440--1448, 2015.

\bibitem{Gysel2016HardwareorientedAO}
Philipp Gysel, Mohammad Motamedi, and Soheil Ghiasi.
\newblock Hardware-oriented approximation of convolutional neural networks.
\newblock {\em CoRR}, abs/1604.03168, 2016.

\bibitem{Han2015DeepCC}
Song Han, Huizi Mao, and William~J. Dally.
\newblock Deep compression: Compressing deep neural network with pruning,
  trained quantization and huffman coding.
\newblock {\em CoRR}, abs/1510.00149, 2015.

\bibitem{Han2015LearningBW}
Song Han, Jeff Pool, John Tran, and William~J. Dally.
\newblock Learning both weights and connections for efficient neural network.
\newblock In {\em NIPS}, 2015.

\bibitem{hinton2015distilling}
Geoffrey Hinton, Oriol Vinyals, and Jeff Dean.
\newblock Distilling the knowledge in a neural network.
\newblock {\em arXiv preprint arXiv:1503.02531}, 2015.

\bibitem{Hinton2015DistillingTK}
Geoffrey~E. Hinton, Oriol Vinyals, and Jeffrey Dean.
\newblock Distilling the knowledge in a neural network.
\newblock {\em CoRR}, abs/1503.02531, 2015.

\bibitem{hoiem2012diagnosing}
Derek Hoiem, Yodsawalai Chodpathumwan, and Qieyun Dai.
\newblock Diagnosing error in object detectors.
\newblock In {\em European conference on computer vision}, pages 340--353.
  Springer, 2012.

\bibitem{huang2017like}
Zehao Huang and Naiyan Wang.
\newblock Like what you like: Knowledge distill via neuron selectivity
  transfer.
\newblock {\em arXiv preprint arXiv:1707.01219}, 2017.

\bibitem{li2016pruning}
Hao Li, Asim Kadav, Igor Durdanovic, Hanan Samet, and Hans~Peter Graf.
\newblock Pruning filters for efficient convnets.
\newblock {\em arXiv preprint arXiv:1608.08710}, 2016.

\bibitem{li2017mimicking}
Quanquan Li, Shengying Jin, and Junjie Yan.
\newblock Mimicking very efficient network for object detection.
\newblock In {\em 2017 IEEE Conference on Computer Vision and Pattern
  Recognition (CVPR)}, pages 7341--7349. IEEE, 2017.

\bibitem{lin2017feature}
Tsung-Yi Lin, Piotr Doll{\'a}r, Ross~B Girshick, Kaiming He, Bharath Hariharan,
  and Serge~J Belongie.
\newblock Feature pyramid networks for object detection.
\newblock In {\em CVPR}, volume~1, page~4, 2017.

\bibitem{lin2018focal}
Tsung-Yi Lin, Priyal Goyal, Ross Girshick, Kaiming He, and Piotr Doll{\'a}r.
\newblock Focal loss for dense object detection.
\newblock {\em IEEE transactions on pattern analysis and machine intelligence},
  2018.

\bibitem{lin2014microsoft}
Tsung-Yi Lin, Michael Maire, Serge Belongie, James Hays, Pietro Perona, Deva
  Ramanan, Piotr Doll{\'a}r, and C~Lawrence Zitnick.
\newblock Microsoft coco: Common objects in context.
\newblock In {\em European conference on computer vision}, pages 740--755.
  Springer, 2014.

\bibitem{liu2016ssd}
Wei Liu, Dragomir Anguelov, Dumitru Erhan, Christian Szegedy, Scott Reed,
  Cheng-Yang Fu, and Alexander~C Berg.
\newblock Ssd: Single shot multibox detector.
\newblock In {\em European conference on computer vision}, pages 21--37.
  Springer, 2016.

\bibitem{luo2017thinet}
Jian-Hao Luo, Jianxin Wu, and Weiyao Lin.
\newblock Thinet: A filter level pruning method for deep neural network
  compression.
\newblock {\em arXiv preprint arXiv:1707.06342}, 2017.

\bibitem{mao2017can}
Jiayuan Mao, Tete Xiao, Yuning Jiang, and Zhimin Cao.
\newblock What can help pedestrian detection?
\newblock In {\em The IEEE Conference on Computer Vision and Pattern
  Recognition (CVPR)}, volume~1, page~3, 2017.

\bibitem{park2016faster}
Jongsoo Park, Sheng Li, Wei Wen, Ping Tak~Peter Tang, Hai Li, Yiran Chen, and
  Pradeep Dubey.
\newblock Faster cnns with direct sparse convolutions and guided pruning.
\newblock {\em arXiv preprint arXiv:1608.01409}, 2016.

\bibitem{rastegari2016xnor}
Mohammad Rastegari, Vicente Ordonez, Joseph Redmon, and Ali Farhadi.
\newblock Xnor-net: Imagenet classification using binary convolutional neural
  networks.
\newblock In {\em European Conference on Computer Vision}, pages 525--542.
  Springer, 2016.

\bibitem{redmon2016you}
Joseph Redmon, Santosh Divvala, Ross Girshick, and Ali Farhadi.
\newblock You only look once: Unified, real-time object detection.
\newblock In {\em Proceedings of the IEEE conference on computer vision and
  pattern recognition}, pages 779--788, 2016.

\bibitem{redmon2017yolo9000}
Joseph Redmon and Ali Farhadi.
\newblock Yolo9000: better, faster, stronger.
\newblock {\em arXiv preprint}, 2017.

\bibitem{ren2015faster}
Shaoqing Ren, Kaiming He, Ross Girshick, and Jian Sun.
\newblock Faster r-cnn: Towards real-time object detection with region proposal
  networks.
\newblock In {\em Advances in neural information processing systems}, pages
  91--99, 2015.

\bibitem{romero2014fitnets}
Adriana Romero, Nicolas Ballas, Samira~Ebrahimi Kahou, Antoine Chassang, Carlo
  Gatta, and Yoshua Bengio.
\newblock Fitnets: Hints for thin deep nets.
\newblock {\em arXiv preprint arXiv:1412.6550}, 2014.

\bibitem{tai2015convolutional}
Cheng Tai, Tong Xiao, Yi Zhang, Xiaogang Wang, et~al.
\newblock Convolutional neural networks with low-rank regularization.
\newblock {\em arXiv preprint arXiv:1511.06067}, 2015.

\bibitem{wen2016learning}
Wei Wen, Chunpeng Wu, Yandan Wang, Yiran Chen, and Hai Li.
\newblock Learning structured sparsity in deep neural networks.
\newblock In {\em Advances in Neural Information Processing Systems}, pages
  2074--2082, 2016.

\bibitem{wen2017coordinating}
Wei Wen, Cong Xu, Chunpeng Wu, Yandan Wang, Yiran Chen, and Hai Li.
\newblock Coordinating filters for faster deep neural networks.
\newblock {\em CoRR, abs/1703.09746}, 2017.

\bibitem{Wu2016QuantizedCN}
Jiaxiang Wu, Cong Leng, Yuhang Wang, Qinghao Hu, and Jian Cheng.
\newblock Quantized convolutional neural networks for mobile devices.
\newblock {\em 2016 IEEE Conference on Computer Vision and Pattern Recognition
  (CVPR)}, pages 4820--4828, 2016.

\bibitem{wu2016quantized}
Jiaxiang Wu, Cong Leng, Yuhang Wang, Qinghao Hu, and Jian Cheng.
\newblock Quantized convolutional neural networks for mobile devices.
\newblock In {\em Proceedings of the IEEE Conference on Computer Vision and
  Pattern Recognition}, pages 4820--4828, 2016.

\bibitem{zagoruyko2016paying}
Sergey Zagoruyko and Nikos Komodakis.
\newblock Paying more attention to attention: Improving the performance of
  convolutional neural networks via attention transfer.
\newblock {\em arXiv preprint arXiv:1612.03928}, 2016.

\bibitem{shufflenet}
Xiangyu Zhang, Xinyu Zhou, Mengxiao Lin, and Jian Sun.
\newblock Shufflenet: An extremely efficient convolutional neural network for
  mobile devices.
\newblock {\em CoRR}, abs/1707.01083, 2017.

\bibitem{zhou2017incremental}
Aojun Zhou, Anbang Yao, Yiwen Guo, Lin Xu, and Yurong Chen.
\newblock Incremental network quantization: Towards lossless cnns with
  low-precision weights.
\newblock {\em arXiv preprint arXiv:1702.03044}, 2017.

\end{thebibliography}
}

\end{document}